# Generalized Hybrid Grey Relation Method for Multiple Attribute Mixed Type Decision Making*

Gol Kim[a], Yunchol Jong[a], Sifeng Liu[b]

[a] Center of Natural Science, University of Sciences, Pyongyang, DPR Korea
[b] College of Economics and Management, Nanjing University of Aeronautics and Astronautics, Nanjing, 210016, China, E-mail: sfliu@nuaa.edu.cn

22 June 2012

## Abstract

The multiple attribute mixed type decision making is performed by four methods, that is, the relative approach degree of grey TOPSIS method, the relative approach degree of grey incidence , the relative membership degree of grey incidence and the grey relation relative approach degree method using the maximum entropy estimation, respectively. In these decision making methods, the grey incidence degree in four-dimensional Euclidean space is used. The final arrangement result is obtained by weighted Borda method. An example illustrates the applicability of the proposed approach.

## 1. Introduction

Ding Chuan-ming, et al (2007) defined a kind of similarity degree for various attribute type and normalized the similarity degree of attribute value of each type in unified metric space. Then, the comparison of between each plan and ideal plan is made by using the new similarity degree and a decision making method was given. Luo Dang, Liu Si-feng (2005) studied, on the basis of the grey relation decision-making, the grey
relation relative approach degree using the maximum entropy estimation. Dang Luo, Li Sun, Sifeng Liu (2010), based on the characteristics of three-parameter interval grey numbers and the advantages of grey target, developed multi-objective grey-target decision making methods under the information of three-parameter interval grey numbers when the grey target weight is known or is unknown. Chunqiao Tan (2011) studied an extension of TOPSIS to a group decision environment, where interdependent or interactive characteristics among criteria and preference of decision makers are taken into account. Some operational laws on interval-valued intuitionistic fuzzy values are introduced and then a generalized interval-valued intuitionistic fuzzy geometric aggregation operator is proposed. Ting-Yu Chen (2012) takes the simple additive weighting (SAW) method and the TOPSIS as the main structure to deal with interval-valued fuzzy evaluation information and presents SAW-based and TOPSIS-based MCDA (multiple-criteria decision analysis) methods. Gong Yanbing, Zhang Jiguo and Deng Jiangao (2008) studied the interval multi-attribute decision making (IMADM) problems, in which the information about attribute weights is known partly and the decision maker has preference information on alternatives in the form of interval numbers reciprocal judgment matrix. An extended TOPSIS method for group decision making with Atanassov's interval-valued intuitionistic fuzzy numbers is proposed to solve the supplier selection

(* This work was supported in part by Nanjing University of Aeronautics and Astronautics, China)

problem under incomplete and uncertain information environment (Gui-Wu Wei, 2011). Yan Chi, Dong-hong Wang (2012) determined the optimal alternative by the shortest distance from the 2-tuple linguistic positive ideal solution (TLPIS) and on the other side the farthest distance of the 2-tuple linguistic negative ideal solution (TLNIS). When the input for a decision process is linguistic, it can be understood that the output should also be linguistic. For that reason, Elio Cables , M. Socorro García-Cascales , M. Teresa Lamata (2012) proposes a modification of the TOPSIS algorithm which develops the above idea and which can also be used as a linguistic classifier. Yahia Zare Mehrjerdi (2012) presents an effective fuzzy multi-criteria method based upon the fuzzy model and the concepts of positive ideal and negative ideal solution points for prioritizing alternatives using inputs from a team of decision makers. For the problem of MADM, in which the attribute values are the interval-valued trapezoidal intuitionistic fuzzy numbers and the weights of attributes are intervals, WAN Shu-ping (2012) proposed a decision making method based on fractional programming, in which Hamming and Euclidean distances for interval-valued trapezoidal intuitionistic fuzzy numbers are defined and by TOPSIS, the models of nonlinear fractional programming for alternative's relative closeness are built.

In all of these papers, the hybrid grey relation decision making methods using the grey incidence degree in four-dimensional Euclidean space have not been discussed yet.

In this paper, first, we present a new method of determining comprehensive weight as grey interval number by considering totally the subjective and the objective weight. Then, we solve MADM problem with various mixed types of attribute by four plan evaluation methods using grey incidence degree in four-dimensional Euclidean space. The methods are (i) the evaluation by the relative approach degree of grey TOPSIS, (ii) the evaluation by the relative approach degree of grey incidence, (iii) the evaluation by the relative membership degree of grey incidence and (iv) the evaluation by the grey relation relative approach degree using the maximum entropy estimation. The final rank is determined by the weighted Borda method using ranks obtained in the above four methods. An application example shows that our method is scientific and practical.

## 2. Decision making problem with mixed type multiple attribute

Let $P = \{p_1, p_2, \cdots, p_n\}$ be a set of plans, $A = \{A_1, A_2, \cdots, A_m\}$ a set of attributes and $T = \{T_1, T_2, T_3, T_4\}$ = {real number, interval real number, linguistic value, uncertain linguistic value} a set of attribute types.

[**Definition 1**] Let $\tilde{u} = S(a)$ be linguistic value, where $S(a)$ is a linguistic measure. They are given by $S = \{S(-5), S(-4), \cdots, S(4), S(5)\}$ = { extremely low, very low, low, comparatively low, a little low, general, a little high, comparatively high, high, very high, extremely high} and $a = \{-5, -4, \cdots, 4, 5\}$.

Supposing that $S(-5) \prec S(-4) \prec \cdots \prec S(4) \prec S(5)$, if $S(i_1) \prec S(i_2) \prec \cdots \prec S(i_n)$, then $\max\{S(i_1), S(i_2), \cdots, S(i_n)\} = S(i_n)$ and $\min\{S(i_1), S(i_2), \cdots, S(i_n)\} = S(i_1)$.

Each linguistic value can be represented by triangle fuzzy number $S = [a^L, a^M, a^U]$, $a^L \leq a^M \leq a^U$, which has a membership function defined by

$$\mu_S(x) = \begin{cases} (x-a^L)/(a^M-a^L), & a^L \leq x \leq a^M \\ (x-a^L)/(a^M-a^L), & a^L \leq a^M \leq a^L \\ 0, & \text{otherwise} \end{cases}.$$

The expression forms of triangle fuzzy number corresponding to $S$ are as follows.
'extremely low' = [0, 0, 0.1], 'very low'= [0, 0.1, 0.2], 'low'= [0.1, 0.2, 0.3], 'comparatively low'= [0.2, 0.3, 0.4], 'a little low' = [0.3, 0.4, 0.5], 'ordinary' = [0.4, 0.5, 0.6], 'a little

high'=[0.5, 0.6, 0.7], 'comparatively high' = [0.6, 0.7, 0.8], 'high' = [0.7, 0.8, 0.9], 'very high'= [0.8, 0.9, 1.0], 'extremely high' = [0.9, 1.0, 1.0].

[**Definition 2**] Let $\tilde{A}=[\alpha,\beta,\gamma,\delta]$ be trapezoid fuzzy number. Then, its membership function is defined by

$$\mu_{\tilde{A}}(x) = \begin{cases} (x-\alpha)/(\beta-\alpha), & \alpha \leq x \leq \beta \\ 1, & \beta < x < \gamma \\ (\delta-x)/(\delta-\gamma), & \gamma \leq x \leq \delta \\ 0, & \text{otherwise} \end{cases}$$

Let $\tilde{A}=[\alpha_1,\beta_1,\gamma_1,\delta_1]$ and $\tilde{B}=[\alpha_2,\beta_2,\gamma_2,\delta_2]$ be two trapezoid fuzzy numbers. Then, the operation rules of trapezoid fuzzy numbers are defined by

$$\tilde{A} \oplus \tilde{B} = [\alpha_1+\alpha_2, \beta_1+\beta_2, \gamma_1+\gamma_2, \delta_1+\delta_2],$$
$$\tilde{A} \otimes \tilde{B} = [\alpha_1\alpha_2, \beta_1\beta_2, \gamma_1\gamma_2, \delta_1\delta_2], \quad k\tilde{A} \otimes \tilde{B} = [k\alpha_1, k\beta_1, k\gamma_1, k\delta_1].$$

[**Definition 3**] Let $S^L = [a^L, a^M, a^U]$ and $S^U = [b^L, b^M, b^U]$. A trapezoid fuzzy number $\tilde{\mu} = (a^L, a^M, b^M, b^U)$ defined by the membership function such as

$$\mu(x) = \begin{cases} (x-a^L)/(a^M-a^L), & a^L \leq x \leq a^M \\ 1, & a^M \leq x \leq b^M \\ (x-b^U)/(b^M-b^L), & b^M \leq x \leq b^U \\ 0, & \text{otherwise} \end{cases}$$

is called a uncertain linguistic value with lower bound $S^L$ and upper bound $S^U$.

[**Definition 4**] Let $a_{ij} = (a_{ij}^{(1)}, a_{ij}^{(2)}, a_{ij}^{(3)}, a_{ij}^{(4)})$ ($a_{ij}^{(1)} \leq a_{ij}^{(2)} \leq a_{ij}^{(3)} \leq a_{ij}^{(4)}$). Then $a_{ij}$ is called a generalized attribute value of $i$ plan for $j$ attribute.

The concrete types of $a = (a^{(1)}, a^{(2)}, a^{(3)}, a^{(4)})$ are such as

- real number type: $a^{(1)} = a^{(2)} = a^{(3)} = a^{(4)}$;
- interval real number type: $a^{(1)} = a^{(2)} < a^{(3)} = a^{(4)}$;
- linguistic value type: $a^{(1)} < a^{(2)} = a^{(3)} < a^{(4)}$;
- uncertain linguistic value type: $a^{(1)} < a^{(2)} < a^{(3)} < a^{(4)}$).

[**Definition 5**] Let $a = (a^{(1)}, a^{(2)}, a^{(3)}, a^{(4)})$ and $b = (b^{(1)}, b^{(2)}, b^{(3)}, b^{(4)})$ be the generalized attribute values, respectively. The distance of between $a$ and $b$ is defined by $d(a,b) = \sqrt{(b^{(1)}-a^{(1)})^2 + (b^{(2)}-a^{(2)})^2 + (b^{(3)}-a^{(3)})^2 + (b^{(4)}-a^{(4)})^2}$.

Let $A = \{A_1, A_2, \cdots, A_n\}$ be a set of plans and $G = \{G_1, G_2, \cdots, G_m\}$ a set of attributes. The attribute value of plan $A_i$ for the attribute $S_j$ is given by the generalized attribute value $a_{ij} = (a_{ij}^{(1)}, a_{ij}^{(2)}, a_{ij}^{(3)}, a_{ij}^{(4)})$ ($i = \overline{1,n}$, $j = \overline{1,m}$).

Let $a_i = \{a_{i1}, a_{i2}, \cdots, a_{im}\}$ be a plan vector and $R = \{a_{ij}\}_{n \times m}$ be a decision matrix. The normalized decision matrix $X = \{x_{ij}\}_{n \times m}$ is obtained by the following method. If $a_{qr}$ is the cost-type attribute, then $x_{qr} = [\underline{x}_{qr}, \overline{x}_{qr}]$ is obtained by the normalization $\underline{x}_{qr} = \dfrac{1/\overline{a}_{qr}}{\sum\limits_{q=1}^{n}(1/\underline{a}_{qr})}$,

$\overline{x}_{qr} = \dfrac{1/\underline{a}_{qr}}{\sum\limits_{q=1}^{n}(1/\overline{a}_{qr})}$. If $a_{qr}$ is of effect type, then $x_{qr}$ is obtained by

$$\underline{x}_{qr} = \frac{\underline{a}_{qr}}{\sum_{q=1}^{n} \overline{a}_{qr}}, \ \overline{x}_{qr} = \frac{\overline{a}_{qr}}{\sum_{q=1}^{n} \underline{a}_{qr}}.$$
If $a_{qr}$ is the triangle fuzzy number such as $a_{qr} = [a_{qr}^L, a_{qr}^*, a_{qr}^U]$,

then $x_{qr} = [x_{qr}^L, x_{qr}^*, x_{qr}^U]$ is obtained by the normalization $x_{qr}^L = \frac{a_{qr}^L}{\sum_{q=1}^{n} a_{qr}^*}$, $x_{qr}^* = \frac{a_{qr}^*}{\sum_{q=1}^{n} a_{qr}^*}$,

$x_{qr}^U = \frac{a_{qr}^U}{\sum_{q=1}^{n} a_{qr}^*}$. If $a_{qr}$ is trapezoid fuzzy number such as $a_{qr} = [a_{qr}^L, a_{qr}^*, a_{qr}^{**}, a_{qr}^U]$,

then $x_{qr} = [x_{qr}^L, x_{qr}^*, x_{qr}^{**}, x_{qr}^U]$ is obtained by normalization such as

$$x_{qr}^L = \frac{a_{qr}^L}{\sum_{q=1}^{n} a_{qr}^*}, \ x_{qr}^* = \frac{a_{qr}^*}{\sum_{q=1}^{n} a_{qr}^*}, \ x_{qr}^{**} = \frac{a_{qr}^{**}}{\sum_{q=1}^{n} a_{qr}^{**}}, \ x_{qr}^U = \frac{a_{qr}^U}{\sum_{q=1}^{n} a_{qr}^{**}}.$$

**[Definition 6]** Let $X = \{x_{ij}\}_{n \times m}$ be a normalized decision matrix. Then $x_i = \{x_{i1}, x_{i2}, \cdots, x_{im}\}$ $(i = \overline{1,n})$ is called an attribute vector of plan $A_i$, where $x_{ij} = (x_{ij}^{(1)}, x_{ij}^{(2)}, x_{ij}^{(3)}, x_{ij}^{(4)})$ is generalized attribute representing the objective preference of decision-maker for the attribute $G_j$.

## 3. Determining of attribute weight

### 3.1. Subjective weight of attribute
The subjective weight of attribute is determined using group AHP method by a decision-making group consisting of $L$ decision-experts.
Let $\alpha_l = [\alpha_l^1, \cdots, \alpha_l^j, \cdots, \alpha_l^m]$ $(l = \overline{1,L})$ be the attribute weight determined by AHP from decision-makers. By using the weights $\alpha_1, \cdots, \alpha_L$ determined by L decision-makers, the weight of attribute $G_j$ is determined by interval grey number $\alpha_j(\otimes)(j = \overline{1,m})$ such as

$$\alpha_j(\otimes) \in [\underline{\alpha}_j, \overline{\alpha}_j], \ 0 \leq \underline{\alpha}_j \leq \overline{\alpha}_j, j = \overline{1,m},$$
where $\underline{\alpha}_j = \min_{1 \leq l \leq L} \{\alpha_l^j\}$, $\overline{\alpha}_j = \max_{1 \leq l \leq L} \{\alpha_l^j\}$, $j = \overline{1,m}$.

### 3.2. Objective weight in the case of generalized attribute values (Optimization)
We define the deviation of decision plan $A_i$ from all other decision plans for attribute $G_j$ in normalized decision matrix $X = (x_{ij}(\otimes))_{n \times m}$ as follows.

$$D_{ij}(\beta^{opt}) = \sum_{k=1}^{m} d(x_{ij}, x_{kj})\beta_j^{opt} =$$

$$= \sum_{k=1}^{m} \sqrt{(x_{kj}^{(1)} - x_{ij}^{(1)})^2 + (x_{kj}^{(2)} - x_{ij}^{(2)})^2 + (x_{kj}^{(3)} - x_{ij}^{(3)})^2 + (x_{kj}^{(4)} - x_{ij}^{(4)})^2} \ \beta_j^{opt}.$$

In order to find proper weight vector $\beta^{opt}$ such that sum of overall deviation for the decision plan attains maximum, we define a deviation function

$$D(\beta) = \sum_{j=1}^{m} \sum_{i=1}^{n} \sum_{k=1}^{n} d(x_{ij}, x_{kj})\beta_j$$

and solve the following nonlinear programming problem.

[P1]  $\max D(\beta) = \sum_{j=1}^{m}\sum_{i=1}^{n}\sum_{k=1}^{n} d(x_{ij}, x_{kj})\beta_j$ ,  s.t. $\sum_{j=1}^{m}\beta_j^2 = 1$, $\beta_j \geq 0$, $j = \overline{1,m}$.

**[Theorem 1]** The solution of problem P1 is given by

$$\overline{\beta}_j = \frac{\sum_{i=1}^{n}\sum_{k=1}^{n} d(x_{ij}, x_{kj})}{\sqrt{\sum_{j=1}^{m}\left[\sum_{i=1}^{n}\sum_{k=1}^{n} d(x_{ij}, x_{kj})\right]^2}}, \quad j = \overline{1,m}.$$

By the normalization of $\overline{\beta}_j$ ($j = \overline{1,m}$), we obtain

$$\beta_j^{opt} = \frac{\sum_{i=1}^{n}\sum_{k=1}^{n} d(x_{ij}, x_{kj})}{\sum_{j=1}^{m}\sum_{i=1}^{n}\sum_{k=1}^{n} d(x_{ij}, x_{kj})}, \quad (j = \overline{1,m}).$$

### 3.3. Objective weight in the case of generalized attribute values (Entropy method)

The entropy weights of the generalized attribute value $x_{ij} = (x_{ij}^{(1)}, x_{ij}^{(2)}, x_{ij}^{(3)}, x_{ij}^{(4)})$ are obtained for each $x_{ij}^{(k)}$ ($k = 1,2,3,4$) as follows. The value $x_{ij}^{(k)}$ ($k = 1,2,3,4$) is normalized by $p_{ij}^{(k)} = \frac{x_{ij}^{(k)}}{\sum_{i=1}^{n} x_{ij}^{(k)}}$ ($i = \overline{1,n}$; $j = \overline{1,m}$) and the entropy value of the $j$ th attribute is calculated by

$E_j^{(k)} = -\frac{1}{\ln n}\sum_{i=1}^{n} p_{ij}^{(k)} \ln p_{ij}^{(k)}$ ( $j = \overline{1,m}$) . In the above formula, if $p_{ij}^{(k)} = 0$ , then we regard that $p_{ij}^{(k)} \ln p_{ij}^{(k)} = 0$. Then deviation coefficient of the $j$ th attribute is calculated by $\eta_j^{(k)} = 1 - E_j^{(k)}$ ( $j = \overline{1,m}$) . Thus, the entropy weight $\beta^{(k)ent} = (\beta_1^{(k)ent}, \beta_2^{(k)ent}, \cdots, \beta_j^{(k)ent}, \cdots, \beta_m^{(k)ent})$ for $x_{ij}^{(k)}$ ($k = 1,2,3,4$) is such as

$$\beta_j^{(k)ent} = \frac{\eta_j^{(k)}}{\sum_{j=1}^{m}\eta_j^{(k)}} = \frac{1 - E_j^{(k)}}{\sum_{j=1}^{m}(1 - E_j^{(k)})} = \frac{1 - E_j^{(k)}}{m - \sum_{j=1}^{m} E_j^{(k)}} \quad (j = \overline{1,m}, k = \overline{1,4}).$$

### 3.4. Comprehensive objective weights

The comprehensive objective weight is determined by the interval grey number
$\beta(\otimes) = (\beta_1(\otimes), \beta_2(\otimes), \cdots, \beta_j(\otimes), \cdots, \beta_m(\otimes))$, where $\beta_j(\otimes) \in [\underline{\beta}_j(\otimes), \overline{\beta}_j(\otimes)]$
and  $\underline{\beta}_j(\otimes) = \min\{\beta_j^{opt}, \beta_j^{(1)ent}, \beta_j^{(2)ent}, \beta_j^{(3)ent}, \beta_j^{(4)ent}\}$,
$\overline{\beta}_j(\otimes) = \max\{\beta_j^{opt}, \beta_j^{(1)ent}, \beta_j^{(2)ent}, \beta_j^{(3)ent}, \beta_j^{(4)ent}\}$.

### 3.5. Final comprehensive weights

In order to consider totally the importance degree among the attributes, the comprehensive weights are determined by the multiplicative composite method such as

$$w_j(\otimes) = \frac{\alpha_j(\otimes) \times \beta_j(\otimes)}{\sum_{j=1}^{m}\alpha_j(\otimes) \times \beta_j(\otimes)} , \quad (j = \overline{1,m})$$

# 4. Some evaluation methods of decision plans

## 4.1. Evaluation of plan by the relative approach degree of grey TOPSIS method

Assume that the subjective preference value of the plan $A_i$ is given by the generalized value $q_i = (q_i^{(1)}, q_i^{(2)}, q_i^{(3)}, q_i^{(4)})$. Let $\tilde{Z} = \{z_{ij}\}_{n \times m}$ be the normalized decision matrix with the subjective preference such as

$$z_{ij} = \left(\frac{1}{2}(q_i^{(1)} + x_{ij}^{(1)}), \frac{1}{2}(q_i^{(2)} + x_{ij}^{(2)}), \frac{1}{2}(q_i^{(3)} + x_{ij}^{(3)}), \frac{1}{2}(q_i^{(4)} + x_{ij}^{(4)})\right).$$

Let $Y = \{y_{ij}\}_{n \times m}$ be the comprehensive weighted decision matrix such as

$$y_{ij} = w_j(\otimes)z_{ij} = (y_{ij}^{(1)}, y_{ij}^{(2)}, y_{ij}^{(3)}, y_{ij}^{(4)}) \quad (i = \overline{1, n}, j = \overline{1, m}).$$

and $y_i = \{y_{i1}, y_{i2}, \cdots, y_{ij}, \cdots, y_{im}\}$ the attribute vector of each plan $(i = \overline{1, n})$.

**[Definition 7]** Let $y_j^+ = \{y_j^{+(1)}, y_j^{+(2)}, y_j^{+(3)}, y_j^{+(4)}\}$, $y_j^{+(k)} = \max_{1 \leq i \leq n}\{y_{ij}^{(k)}\}$ $(k = 1, 2, 3, 4)$.

Then, the $m$-dimensional interval grey number vector $y^+ = \{y_1^+, y_2^+, \cdots, y_m^+\}$ is called a positive ideal plan attribute vector.

Let $y_j^- = \{y_j^{-(1)}, y_j^{-(2)}, y_j^{-(3)}, y_j^{-(4)}\}$, $y_j^{-(k)} = \min_{1 \leq i \leq n}\{y_{ij}^{(k)}\}$ $(k = 1, 2, 3, 4)$. Then, the $m$-dimensional interval grey number vector $y^- = \{y_1^-, y_2^-, \cdots, y_m^-\}$ is called a negative ideal plan attribute vector.

Euclidian distance between each plan attribute vector $y_i$ and the positive ideal plan attribute vector $y^+$ or the negative ideal plan attribute vector $y^-$ is

$$D_i^+ = \sqrt{\sum_{j=1}^{m}\left[(y_{ij}^{(1)} - y_j^{+(1)})^2 + (y_{ij}^{(2)} - y_j^{+(2)})^2 + (y_{ij}^{(3)} - y_j^{+(3)})^2 + (y_{ij}^{(4)} - y_j^{+(4)})^2\right]}$$

or

$$D_i^- = \sqrt{\sum_{j=1}^{m}\left[(y_{ij}^{(1)} - y_j^{-(1)})^2 + (y_{ij}^{(2)} - y_j^{-(2)})^2 + (y_{ij}^{(3)} - y_j^{-(3)})^2 + (y_{ij}^{(4)} - y_j^{-(4)})^2\right]}$$

The relative approach degree between each plan and the ideal plan is given by

$$C_i = \frac{D_i^-}{D_i^+ + D_i^-} \quad (i = \overline{1, n}) \quad .$$

The best plan corresponds to the largest $C_i$.

## 4.2. Evaluation of plan by the relative approach degree of grey incidence

**[Definition 8]** Let $Y = \{y_{ij}\}_{n \times m}$ be the normalized comprehensive weighted decision matrix and $y_j^+$ ($y_j^-$) be the positive (negative) ideal plan attribute vector. We define

$$r_{ij}^+ = \frac{\min_i \min_j d(y_{ij}, y_j^+) + \rho \max_i \max_j d(y_{ij}, y_j^+)}{d(y_{ij}, y_j^+) + \max_i \max_j d(y_{ij}, y_j^+)},$$

$$r_{ij}^- = \frac{\min_i \min_j d(y_{ij}, y_j^-) + \rho \max_i \max_j d(y_{ij}, y_j^-)}{d(y_{ij}, y_j^-) + \max_i \max_j d(y_{ij}, y_j^-)}.$$

Then, $r_{ij}^+$ ($r_{ij}^-$) is called a grey incidence coefficient of the positive ideal factor $y_j^+$ (the negative ideal factor $y_j^-$) with respect to the factor $y_{ij}$, where $\rho \in (0, 1)$ and $\rho = 0.5$ is taken in general.

**[Definition 9]** Matrix $P^+ = \{r_{ij}^+\}_{n\times m}$ ( $P^- = \{r_{ij}^-\}_{n\times m}$) is called a grey incidence coefficient matrix of the given plan with respect to the positive (negative) ideal plan.
We define

$$G(y^+, y_i) = \frac{1}{m}\sum_{j=1}^{m} r_{ij}^+ \ (i=\overline{1,n}), \qquad G(y^-, y_i) = \frac{1}{m}\sum_{j=1}^{m} r_{ij}^- \ (i=\overline{1,n})$$

Then matrix $G(y^+, y_i)$ ($G(y^-, y_i)$) is called a grey interval incidence degree of the comprehensive attribute vector for the plan $A_i$ with respect to the positive(negative) ideal plan attribute vector.

**[Theorem 2]** Let $\overline{Y} = \{y_0, y_1, \cdots, y_m\}$ be the set of grey relation factors, where $y_0 = y^+$. Then the grey interval incidence degrees $G(y_i^+, y_j)$ and $G(y^-, y_i)$ satisfy four axioms of the grey incidence degree, i.e. normality, pair-symmetry, wholeness and closeness.

This is different from the relative approach degree in traditional TOPSIS.
The degree of grey incidence relative approach is modified by introducing the preference coefficient as follows.

$$C_i = \begin{cases} \dfrac{G(y^+, y_i)\cdot\theta_+}{G(y^+, y_i)\cdot\theta_+ + G(y^-, y_i)\cdot\theta_-} ; & 0<\theta_+ <1, \theta_- <1 \\ G(y^+, y_i) ; & \theta_+ = 1, \theta_- = 0 \end{cases}$$

where $\theta_+$ and $\theta_-$ are the preference coefficients, respectively. $\theta_+$ reflects the degree of preference or degree of interest of the evaluation plan with respect to the positive ideal plan. $\theta_-$ reflect the degree of preference or degree of interest of the evaluation plan with respect to the negative ideal plan. Generally, we regard as $\theta_+ > \theta_-$ and choose it so as to satisfy $0 < \theta_+ \leq 1$, $0 < \theta_- \leq 1$, $\theta_+ + \theta_- = 1$.
The best plan corresponds to the largest value of the grey incidence relative approach degree $C_i$.

### 4.3. Evaluation of plan by the relative membership degree of grey incidence

If the membership degree of the positive ideal plan with respected the plan $A_i$ is $u_i$, from the definition of rest sets of fuzzy set theory, the membership degree of the negative ideal plan corresponding to the plan $A_i$ is $1-u_i$. Therefore, we can find the membership degree vector $u = (u_1, u_2, \cdots, u_n)$ by solving the following problem.

[P2] $\qquad \min F(u) = \sum_{i=1}^{n}\{[(1-u_i)G(y^+, y_i)]^2 + [u_i G(y^+, y_i)]^2\}$.

**[Theorem 3]** The optimal solution of the optimization problem P2 is given by

$$u_i = \frac{G^2(y^+, y_i)}{G^2(y^+, y_i) + G^2(y^-, y_i)} \ (i=\overline{1,n}).$$

The best plan is one with the largest value of $u_i$.

### 4.4. Evaluation of plan by the grey relation relative approach degree using maximum entropy estimation

**[Definition 10]** Let $G(y^+, y_i)$ and $G(y^-, y_i)$ be the grey interval incidence degree for the plan $A_i$ with respect to the positive ideal plan and the negative ideal plan, respectively. We denote the weights of these two grey interval incidence degrees by $\beta_1$ and $\beta_2$ ($\beta_1 + \beta_2 = 1$, $\beta_1, \beta_2 \geq 0$), respectively. Then,

$$C_i^{''} = \beta_1 G(y^+, y_i) + \beta_2[1 - G(y^-, y_i)] \ (i = \overline{1,n})$$

is called a grey comprehensive incidence degree of the factor vector $y_i$.

To obtain $\beta_1$ and $\beta_2$ by entropy method, we solve the following optimization problem

[P3] $\quad \max\{\sum_{i=1}^{n}[\beta_1 G(y^+, y_i) + \beta_2(1 - G(y^-, y_i))] - \sum_{j=1}^{2} \beta_j \ln \beta_j\}$

$$s.t. \begin{cases} \beta_1 + \beta_2 = 1, \\ \beta_1 \geq 0, \beta_2 \geq 0 \end{cases}.$$

By solving this problem, we obtain the following weights.

$$\beta_1 = e^{\sum_{i=1}^{n}(G(y^+, y_i) + G(y^-, y_i) - 1)} (1 + e^{\sum_{i=1}^{n}(G(y^+, y_i) + G(y^-, y_i))})^{-1},$$

$$\beta_2 = (1 + e^{\sum_{i=1}^{n}(G(y^+, y_i) + G(y^-, y_i))})^{-1}.$$

The best plan is one with the largest value of $C_i^*$.

The final rank is determined by the weighted Borda method using rank vectors obtained in the above four methods.

## 5. An illustrative example

Let's consider the decision-making problem for five programs of fighter development. The decision matrix is given in Table 1, in which $G_1, G_2, G_3, G_4, G_5$ denote plans and $A_1, A_2, \cdots, A_9$ denote attributes (index). The meaning of attributes is such as; $A_1$ - weight empty of body(Kg), $A_2$ - flight radius(Km), $A_3$ - maximum flying speed(Km/h), $A_4$ - development cost (ten thousand Yuan), $A_5$ - reversal of body head(h), $A_6$ - maintenance possibility, $A_7$ - security, $A_8$ - reliability level of development group, $A_9$ - degree of environmental influence.

**Table 1.** Decision matrix

| Index<br>Plan | $A_1$ | $A_2$ | $A_3$ | $A_4$ | $A_5$ |
|---|---|---|---|---|---|
| $G_1$ | 3610 | 490 | [465, 485] | [4830, 4910] | [850, 950] |
| $G_2$ | 3590 | 520 | [480, 490] | [4680, 4790] | [800, 900] |
| $G_3$ | 3700 | 480 | [465, 475] | [4600, 4720] | [700, 800] |
| $G_4$ | 3780 | 470 | [460, 475] | [4660, 4770] | [700, 750] |
| $G_5$ | 3690 | 510 | [470, 485] | [4770, 4850] | [750, 850] |

**Table 1.** Decision matrix (continued)

| Index<br>Plan | $A_6$ | $A_7$ | $A_8$ | $A_9$ |
|---|---|---|---|---|
| $G_1$ | very high | rather high | [a little high, rather high] | [low, a little low] |
| $G_2$ | high | high | [high, a little high] | [low, rather low] |
| $G_3$ | rather high | rather high | [rather high, very high] | [very low, rather low] |
| $G_4$ | general | a little high | [general, very high] | [rather low, a little low] |
| $G_5$ | rather high | high | [rather high, high] | [very low, Low] |

We assume that two experts are invited to determine the subjective attribute weights by AHP method. Thus, the subjective weight obtained from group AHP method is

$\alpha(\otimes) = ([\underline{\alpha}_1, \overline{\alpha}_1], [\underline{\alpha}_2, \overline{\alpha}_2], [\underline{\alpha}_3, \overline{\alpha}_3], [\underline{\alpha}_4, \overline{\alpha}_4], [\underline{\alpha}_5, \overline{\alpha}_5], [\underline{\alpha}_6, \overline{\alpha}_6], [\underline{\alpha}_7, \overline{\alpha}_7], [\underline{\alpha}_8, \overline{\alpha}_8], [\underline{\alpha}_9, \overline{\alpha}_9])$

= ([0.2305, 0.3093], [0.1501, 0.1675], [0.1262, 0.1761], [0.1323, 0.1348], [0.0815, 0.0948], [0.0557, 0.0622], [0.0431, 0.0623], [0.0492, 0.0515], [0.0352, 0.0376])

The subjective preference values of decision-making group to plans are such as

$q_1$ = (0.2, 0.3, 0.3, 0.4), $q_2$ = (0.2, 0.4, 0.4, 0.5), $q_3$ = (0.1, 0.2, 0.3, 0.4),
$q_4$ = (0.1, 0.2, 0.3, 0.4), $q_5$ = (0.2, 0.3, 0.4, 0.5).

The relative approach degree of grey TOPSIS method is $C$ = (0.5588, 0.9845, 0.1913, 0.3674, 0.6086). Thus, we obtain the rank such as

Plan 2 $\succ$ plans 5 $\succ$ plan 1 $\succ$ plan 4 $\succ$ plan 3

Then, we calculate the degree of grey relation relative approach with the preference coefficients. We take the preference coefficients $\theta_+ = \theta_- = 0.5$. The obtained $C'$ is such as

$C'$ = (0.4688, 0.6050, 0.4222, 0.4203, 0.5602). Thus, we obtain the rank of plans such as

Plan 2 $\succ$ plans 5 $\succ$ plan 1 $\succ$ plan 3 $\succ$ plan 4.

Finding the relative membership degree between the given plans and the positive ideal plan, we obtain $u = (u_1, u_2, u_3, u_4, u_5)$ = (0.4379, 0.7011, 0.3480, 0.3445, 0.6187) and the rank such as

Plan 2 $\succ$ plans 5 $\succ$ plan 1 $\succ$ plan 3 $\succ$ plan 4.

The grey incidence relative approach degree by using the maximum entropy is $C''$ = (0.7266, 0.9735, 0.6864, 0.6891, 0.8656). Thus, we obtain the rank such as

Plan 2 $\succ$ plans 5 $\succ$ plan 1 $\succ$ plan 4 $\succ$ plan 3

The final rank by the weighted Borda method is such as.

Plan 2 $\succ$ plans 5 $\succ$ plan 1 $\succ$ plan 3 $\succ$ plan 4.

## 6. Conclusion

In this paper, by introducing four-dimensional Euclidean distance for the hybrid MADM problems, the different types of attribute are normalized in a unified metric space and it has a unified quantification. We have proposed a new grey relation interval decision-making method for MADM problem in which the attribute value is given by various mixed-type such as real, interval, linguistic and uncertain linguistic value. The first feature of our method is to find the subjective and objective weights, based on group AHP, and optimization and entropy method, respectively, as interval numbers and to find the interval final weights by multiplicative composite rule so as to consider totally the subjectiveness and the objectiveness in determining weights. The second feature is to introduce grey interval weight in decision matrix of mixed multiple attribute so that each element of the decision matrix becomes an element of four-dimensional Euclidean space, and to propose four methods such as the evaluation by the relative approach degree of grey TOPSIS, the evaluation by the relative approach degree of grey incidence, the evaluation by the relative membership degree of grey incidence and the evaluation by the grey relation relative approach degree using the maximum entropy estimation for ranking of plans, and the determine the final rank by weighted Borda method using the rank vectors of the above four methods. Thus, our method can make the decision-making more scientific and reliable.